\documentclass[10pt,twocolumn,letterpaper]{article}
\usepackage{cvpr}
\usepackage{times}
\usepackage{epstopdf}
\usepackage{hyperref}       
\usepackage{url}            
\usepackage{booktabs}       
\usepackage{amsfonts}       
\usepackage{nicefrac}       
\usepackage{microtype}      
\usepackage{multirow}
\usepackage{epsfig}
\usepackage{graphicx}
\usepackage{amsmath,amssymb}
\usepackage[nolist]{acronym}

\cvprfinalcopy 
\def\cvprPaperID{4019} 

\begin{acronym}
\acrodef{cnn}[CNN]{Convolutional Neural Network}
\acrodef{dl}[DL]{Deep Learning}
\acrodef{hpo}[HPO]{Hyperparameter Optimization}
\acrodef{ml}[ML]{Machine Learning}
\acrodef{rnn}[RNN]{Recurrent Neural Network}
\acrodef{cgp}[CGP]{Cartesian Genetic Programming}
\acrodef{mcts}[MCTS]{Monte Carlo Tree Search}
\acrodef{smbo}[MSBO]{Sequential Model-Based Optimization}
\acrodef{gpu}[GPU]{Graphics Processing Unit}
\end{acronym}

\begin{document}

\title{Effective Building Block Design for \\Deep Convolutional Neural Networks using Search}

%



\author{Jayanta K Dutta\textsuperscript{1}, Jiayi Liu\textsuperscript{1},
	Unmesh Kurup\textsuperscript{1},
	Mohak Shah\textsuperscript{2}\\
	\textsuperscript{1}{Bosch Center for Artificial Intelligence, Sunnyvale, CA, USA}\\
	\textsuperscript{2}{University of Illinois at Chicago, Chicago, Illinois, USA} \\
	\texttt{\{Jayanta.Dutta, Jiayi.Liu2, Unmesh.Kurup\}@us.bosch.com, mohak@mohakshah.com}
}


\maketitle

\begin{abstract}

Deep learning has shown promising results on many machine learning tasks but DL models are often complex networks with large number of neurons and layers, and recently, complex layer structures known as building blocks. Finding the best deep model requires a combination of finding both the right architecture and the correct set of parameters appropriate for that architecture. In addition, this complexity (in terms of layer types, number of neurons, and number of layers) also present problems with generalization since larger networks are easier to overfit to the data. In this paper, we propose a search framework for finding effective architectural building blocks for convolutional neural networks (CNN). Our approach is much faster at finding models that are close to state-of-the-art in performance. In addition, the models discovered by our approach are also smaller than models discovered by similar techniques. We achieve these twin advantages by designing our search space in such a way that it searches over a reduced set of state-of-the-art building blocks for CNNs including residual block, inception block, inception-residual block, ResNeXt block and many others. We apply this technique to generate models for multiple image datasets and show that these models achieve performance comparable to state-of-the-art (and even surpassing the state-of-the-art in one case). We also show that learned models are transferable between datasets.

\end{abstract}

\section{Introduction}

Deep \acp{cnn} currently produce state-of-the-art accuracy on many machine learning tasks including image classification.  
Early \ac{dl} architectures such as AlexNet \cite{krizhevsky2012imagenet}, ZF net \cite{zeiler2014visualizing}, and VGG net \cite{simonyan2014very} used only convolution, fully connected, and/or pooling operations but still provided large improvements over classical vision approaches. Recent advances in the field have improved performance further by using several new and more complex building blocks that involve operations such as branching (e.g. inception \cite{szegedy2015going}, ResNeXt \cite{xie2016aggregated} blocks) and skip connections (e.g. residual \cite{he2016deep}, ResNeXt \cite{xie2016aggregated}).   
Since the set of operations to be used for each branch remains an active area of research, find the correct building block involves searching over the possible configurations of branch components. This increase in the search space effectively means that in addition to traditional Deep CNN hyperparameters such as layer size and number of filters, training a model now includes searching over the various combinations involved in constructing an effective building block. This increased complexity corresponds to increased training time and often means that the process of finding the right architecture or configuration remains the result of extensive search.

Recently, there has been some research in tackling this issue by automating the architecture discovery process. We can consider these methods as falling into one of two categories. The first set of methods focus on discovering the entire architecture from primary building blocks i.e., convolution layers, pooling layers, fully connected layers etc \cite{zoph2016neural, baker2016designing, real2017large, miikkulainen2017evolving}. The other set of methods focus on building these architectures from the afore-mentioned more complex blocks involving branching and skip connections. The goal with this second set of methods is finding one particular building block \cite{zoph2017learning, zhong2017practical} which is then repeated many times to create the deep architecture. With both approaches techniques such as reinforcement learning \cite{zoph2016neural, baker2016designing, zoph2017learning, zhong2017practical} or evolutionary algorithms \cite{real2017large, miikkulainen2017evolving} are generally used to search through the architecture space. One drawback to using such search techniques is that they are computationally expensive.

In this paper, we consider the second approach - that of designing an effective architectural building block which is then repeated to create a deeper network. Motivated from the ResNet \cite{he2016deep} and Inception \cite{szegedy2015going} module structure, our block design includes branching and skip connections. ResNet \cite{he2016deep} only includes 2D convolution ($n \times n$) operations in their block design, Inception \cite{szegedy2015going} includes 2D convolution as well as row/column convolutions ($n \times 1$ or $1 \times n$ filters), and Xception \cite{chollet2016xception} includes separable depth-wise convolution operations. There are also several techniques in the literature for combining the outputs from different branches including concatenation \cite{szegedy2015going}, adding \cite{xie2016aggregated} and summation with stochastic affine transformation \cite{gastaldi2017shake}. We propose a search framework for finding architectural building blocks for CNNs considering all the well-known operations in the branch along with their combination techniques. We use random search over the search space to generate building blocks and repeat this block multiple times to create a deep network. With only 50 architectures searched, we are able to find an architecture providing comparable performance with respect to state-of-the art models on multiple image recognition datasets including CIFAR-10, CIFAR-100, SVHN, and FER2013. Our approach has the additional advantage that the search process is much simpler (random search) than previous approaches (reinforcement learning and evolutionary techniques) that need many more trials to generate architectures with comparable performance. Finally, the models discovered by our approach are smaller (as measured in terms of number of weight parameters) than models discovered by other architecture discovery methods.

\section{Related work}

Deep \ac{cnn}s have shown promising results on many machine learning tasks including image classification. Starting from AlexNet \cite{krizhevsky2012imagenet} to the more modern ResNet \cite{he2016deep} there have been many architectural changes to improve the performance of the deep \ac{cnn}. These improvements have seen, among other achievements, a drop in the top 5 error rate from 15.3\% to 3.57\% for the ImageNet classification task \cite{russakovsky2015imagenet}. Previous CNNs included convolution layers, pooling layers, and fully connected layers. These layers were then stacked to create a deeper network. Recently, different kinds of layers such as depthwise convolution, separable convolution, and dilated convolution, have been introduced. In addition to stacking the layers on top of each other, skip connections are used to pass the gradients smoothly for deeper models. One disadvantage of these advances is that it has become more difficult to select and optimize the best model for a practical application. As a consequence there has been some interest recently in finding architectures automatically. 

\emph{Zoph, Le \cite{zoph2016neural}}: Recurrent Neural Networks (RNNs) are used to generate model descriptions of the neural network to be trained and the \ac{rnn} is trained with reinforcement learning to maximize the expected accuracy of the generated architectures on a validation dataset. 
The \ac{rnn} predicts filter height, filter width, stride height, stride width, and number of filters for one layer and repeats this layer to construct the CNN. Every prediction is carried out by a softmax classifier and then fed into the next time-step as input. Generally, the process of generating an architecture stops if the number of layers exceeds a certain value. After generating an architecture, a neural network with this architecture is built, trained, and validated with a held-out validation set. The validation performance is used to update the policy of the \ac{rnn} to generate better architectures over time.
This approach involves minimal human intervention and depends on the \ac{rnn} to learn the policy from scratch.
However, this approach also involves further tuning of the hyperparameters of the \ac{rnn} model and needs many samples of architectures to learn a good policy.

\emph{Baker et al. \cite{baker2016designing}}: MetaQNN is a meta-modeling algorithm based on reinforcement learning that automatically generates high performing \ac{cnn} architectures for a given learning task. The learning agent sequentially chooses neural network layers with layer types (e.g. convolutional, pooling, fully connected, termination) and their corresponding hyperparameters via $\epsilon$-greedy policy until it reaches a termination state. The generated architecture is trained and validated with a held-out validation set. The validation performance is used as a reward to update the Q-learning network to generate better architectures over time.
While this work extends the architecture generation ability of \cite{zoph2016neural} to more layer types, it still suffers from similar drawbacks resulting from the use of reinforcement learning which requires many samples to accurately learn a good policy. 

\emph{Real et al. \cite{real2017large}}: This approach uses a simple evolutionary algorithm to automatically discover high performance \ac{cnn} models. The algorithm starts with random \ac{cnn} architecture that usually performs poorly and then progressively improves this architecture while navigating a fairly unrestricted search space. Each evolved model is trained and validated with a held-out validation set and the performance is considered as the individual's quality or fitness score. The scheme uses repeated pairwise competitions of random individuals. Different mutations are used for the reproduction steps and the child architectures are allowed to inherit the parents' weights whenever possible.
Given the limited mutation space for each step, it still requires a significant amount of computing resources to reach the optimal solution.

\emph{Miikkulainen et al. \cite{miikkulainen2017evolving}}: CoDeepNEAT is an automated method for optimizing deep learning architectures through evolution. At first, a population of minimal complexity neural networks are generated and over many generations structure (nodes and edges) are added incrementally through mutation. Each node represents a layer in the deep network and contains a table of real and binary valued hyperparameters. These hyperparameters are mutated through an uniform Gaussian distribution and random bit-flipping, respectively. These hyperparameters determine the type and properties of the layer. Each evolved model is trained and validated with a held-out validation set and the performance is considered as the individual's quality or fitness score.
Because it starts from a relatively complex architecture, it is more efficient than the approaches that begin from a cold start.  However since it limits its search space to a few initial neural networks, it is hard to reach the state-of-the-art performance when compared to other architecture search methods.

\emph{Zoph et al. \cite{zoph2017learning}}:NASNet considers learning an architectural building block rather than learning the full architecture. The learned building block is then repeated many times to create the deep architecture. An \ac{rnn} is used to generate the descriptions of the building block in a manner similar to \cite{zoph2016neural}. For \ac{cnn} models, two types of building blocks are learned: \emph{Normal cell} and \emph{Reduction cell}, where these blocks return same size feature maps and reduced size feature maps respectively. The controller \ac{rnn} is trained using proximal policy optimization to generate better architectural building blocks over time. This approach also generates models that are transferable with the best building blocks for one learning task also showing good performance over other learning tasks as well. However the \ac{rnn} itself still lacks transferability and needs to be retrained for each new problem. Also, because the blocks are learned from scratch, this approach requires many more trials to achieve state-of-the-art performance.


\emph{Suganuma et al. \cite{suganuma2017genetic}}:  \ac{cgp} is used to automatically construct \ac{cnn} architectures for an image classification task. The \ac{cnn} structure and connectivity represented by the \ac{cgp} encoding method are optimized to maximize the validation accuracy. The \ac{cnn} architecture defined by the \ac{cgp} is trained and validated with a held-out validation set and the performance is considered as the quality or fitness score of the architecture. However, CGP creates an additional set of hyperparameters (e.g. mutation rate) that now need to be searched over to find the best model.

\emph{Cai et al. \cite{cai2017reinforcement}}: This paper proposes a reinforcement learning search framework where the action is to grow the network depth or layer width based on the current network architecture while preserving functionality. A shared encoder network is used to learn a low-dimensional representation of the given architecture and a separate actor network generates certain types of network transformation actions. After $T$ steps of transformations the final network along with transferred weights is trained and validated with a held-out validation set. The validation performance is used to update the policy using a policy gradient method.
Because the mutation is to either increase the network depth or the layer width, it can efficiently search through very deep architectures. However this also limits its ability to find a less complex but good-enough architecture.

\emph{Negrinho et al. \cite{negrinho2017deeparchitect}}: DeepArchitect provides a framework for automatically designing and training deep models. It proposes an extensible and modular language that allows the human expert to compactly represent complex search spaces over architectures and their hyperparameters. Random search, Monte Carlo tree search (MCTS), and sequential model-based optimization (SMBO) are used to explore the search space.
However the result of the search is an achitecture that performs much worse than architectures found by current state-of-the-art architecture search methods (see Table~\ref{tab:comparison}).

\section{Methodology}

\subsection{Building block design}

\begin{figure}[t]
	\centering
    \includegraphics[width=0.95\columnwidth]{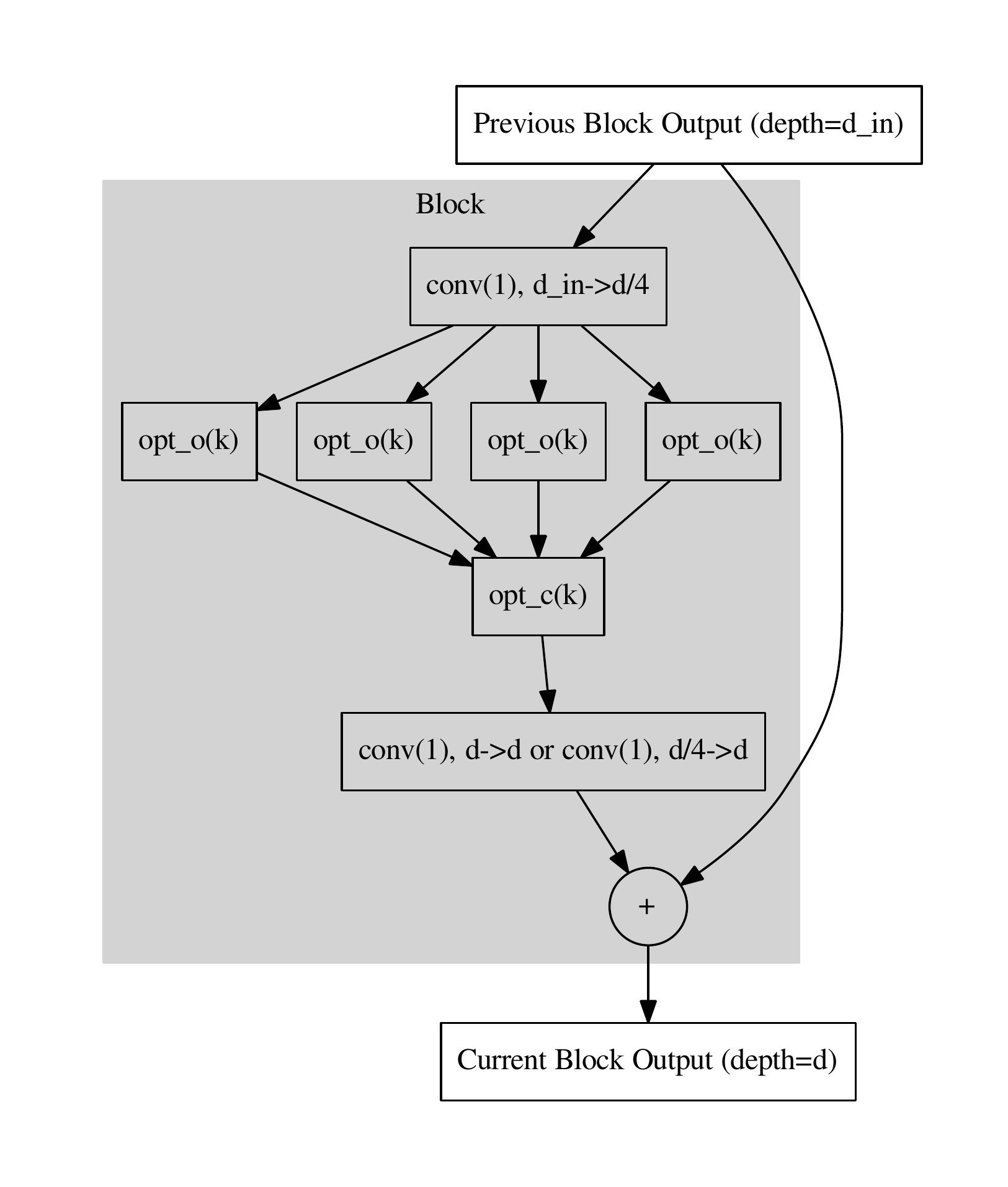}
    \caption{Design of {CNN} building block.}
    \label{fig:s_space}
\end{figure}

The residual building block (ResNet \cite{he2016deep}) has made it easier to train much deeper neural networks while producing state-of-the art results. A residual block simply adds the input of the block to the output of the layer(s) within the block and can be described more formally by 

\begin{equation}
	G(x) = x + F(x),
\end{equation}

\noindent where $x$ is the input of the residual block, $G(x)$ is the output of the residual block and $F(x)$ is the output of a residual branch on the residual block. In the basic design, $F(x)$ contains two $3 \times 3$ convolution layers along with a batch normalization and/or a rectified linear unit activation function. For training deeper nets, the building block is modified as a \emph{bottleneck} design. It contains 3 layers instead of 2 ($1 \times 1$, $3 \times 3$, $1 \times 1$ convolutions) and $1 \times 1$ layers responsible for reducing and then increasing the depth dimension to reduce the number of parameters for deeper nets. 

Several variants of ResNet are available which have multiple branches instead of a single residual branch, e.g. ResNeXt \cite{xie2016aggregated}, Inception-ResNet \cite{szegedy2017inception}, Shake-shake Residual net \cite{gastaldi2017shake}. The ResNext building block performs a set of operations whose outputs are aggregated by summation but all transformations are of the same topology. Inception-ResNet uses the Inception module with skip connections. All the operations and hyperparameters are selected carefully to achieve better performance on the ImageNet classification task. The exact configuration of each module varies throughout the network. Shake-shake Residual nets use two branches in the residual function and the outputs of the branches are combined with the standard summation by a stochastic affine combination. We design our search space to use the same block throughout the network (like ResNext) but allow the operations inside the block to be learned (like Inception-ResNet). This enables us to take combine the functionalities from each of the above models while still limiting the extent of the search space by having a block structure.


Our framework searches through a structure that can be considered as a convolutional cell building block rather than the whole architecture. The building block can then be stacked many times to create a deep CNN. Figure~\ref{fig:s_space} shows the design of our building block. The residual branch starts with a $1 \times 1$ convolutional layer to reduce the feature depth by a factor of 4 with respect to the output feature depth of the block. Unlike ResNet, where it contains a $3 \times 3$ convolutional layer after the bottleneck layer, we create three branches. The operation in each branch can be selected from the following operations with $k \in \{1,3,5\}$:

\begin{itemize}
	\item \emph{conv($k$)}: $k \times k$ convolution which is generally used in designing most CNNs.
	\item {rc\_conv($k$)}: $k \times 1$ convolution followed by $1 \times k$ convolution. Inception-v4 \cite{szegedy2017inception} uses this operation to reduce the number of parameters.
	\item \emph{sp\_conv($k$)}: $k \times k$ depthwise separable convolution consists of a \emph{depthwise convolution} followed by a \emph{pointwise convolution} \cite{chollet2016xception}, which enables more efficient use of model parameters.
\end{itemize}

The outputs from the branches can be combined using one of the following operations:

\begin{itemize}
	\item \emph{concat}: concatenating the outputs in the feature dimension (like in the inception \cite{szegedy2015going} module).
	\item \emph{add\_det}: adding the outputs (like in the ResNeXt \cite{xie2016aggregated} module).
	\item \emph{add\_stc}: adding the outputs, weighted by a random constant (like in the forward pass of shake-shake residual \cite{gastaldi2017shake} block)
\end{itemize}

\noindent Finally a $1 \times 1$ convolution layer is used to increase the feature depth to the output feature depth of the block. Each operation is followed by a batch normalization and rectified linear unit activation function. We use strided operation for doing spatial feature space reduction. In case of feature reduction, $1 \times 1$ convolution with stride 2 is applied on the input feature map of the block to match the dimension of the residual branch before adding them. We double the number of output units in case of spatial feature size reduction to maintain constant hidden state dimension. Each block is repeated $n$ times before any spatial feature size reduction. We consider the number of repetitions $n$ and the number of initial convolution filters as hyperparameters.

\subsection{Search strategy}

As we treat the choices of branches as hyperparameters, we are open to many off-the-shelf optimization methods. Random search is one of the simplest methods for hyperparameter optimization \cite{bergstra2012random}. Compared to iterating over pre-defined parameter combinations (i.e., grid search), random search shows a good balance between exploration and exploitation, and thus better convergence rate. It is also less sensitive to the prior assumptions on the distribution of hyperparameters which makes it a more robust alternative when applied to many different problems. In addition, the random search algorithm is naively parallelizable as there is no dependency on historical results. 

There are of course more advanced methods such as Bayesian optimization, evolutionary algorithms, and reinforcement learning that can be applied to the hyperparameter optimization problem. In the case of exploration vs exploitation techniques, the early stages are explorative and akin to random search \cite{bergstra2013making, young2015optimizing}. Since our approach has a limited search space and uses only a small number of trials (50), a random search is enough to effectively explore this space.  
More recent work \cite{negrinho2017deeparchitect} has also confirmed that at the early stage, random search clearly outperforms other optimization algorithms in the first ten trials. It is not clear that the other methods, e.g. \ac{mcts} and \ac{smbo}, are better given the large uncertainty associated with the training and validation results within the first hundred trials. As the target is to find a better architecture with fewer trials, we choose the random search for the \ac{hpo} process in this work and leave other methods for future investigation.

It would be remiss not to mention that our search strategy is focused only on improving the search space associated with the hyperparameters of the \ac{cnn} architecture. This leaves other model parameters such as the learning rate, momentum, initialization etc to be discovered. Our proposal does not extend to these hyperparameters and techniques to discover them are out of the scope of this study.

\section{Experimental Results}

We experimented on four image classification datasets: CIFAR-10 \cite{krizhevsky2009learning}, CIFAR-100 \cite{krizhevsky2009learning}, SVHN \cite{netzer2011reading}, FER2013 \cite{Fer2013}. CIFAR-10 \cite{krizhevsky2009learning} is an object recognition dataset and has 50000 training examples and 10000 test examples with 10 categories. Among the training examples, 5000 examples were used for validation. CIFAR-100 \cite{krizhevsky2009learning} is also an object recognition dataset, but with 100 categories. It has 50000 training examples and 10000 test examples. Among the training examples, 5000 examples were used for validation. SVHN \cite{netzer2011reading} is a digit recognition dataset and is obtained from house numbers in Google Street View images. It has 73257 training examples and 26032 test examples with 10 categories. Among the training examples, 5000 examples were used for validation. We did not use the additional dataset for training on SVHN dataset. FER2013 \cite{Fer2013} is a facial expression recognition dataset and has 28709 training, 3589 validation and 3589 test examples with 7 categories. We note that the test set in every case is never used for model selection, and only used for final model evaluation.


For all the experiments, we used momentum optimizer to minimize the cross-entropy loss with a minibatch size of 128. The initial learning rate was set to 0.1 and was decreased every 25 epochs by a factor of 0.5. Momentum was set to 0.9 and maximum number of epochs was set to 500. We used early stopping if the validation accuracy stops improving within 50 trials. We used weight decay parameter equal to 0.001 for training and common image preprocessing techniques including per pixel mean subtraction (calculated over the training set), random cropping and random horizontal flipping (except for SVHN dataset). For all the experiments, we selected top-10 models based on validation accuracy and report the test accuracy on the best model. We also construct an ensemble of the top-10 mdoels by averaging their responses. We used the open source project {\em hof} and Tensorflow \cite{abadi2016tensorflow} to carry out the experiments. 

We randomly searched over the operations (\emph{opt\_o}) and their combination types (\emph{opt\_c}) to create a building block. The created block is stacked repeatedly to assemble the deep learning architecture that is then trained. We report results with only 50 architectures searched. 


\begin{figure}[h]
\includegraphics[width=0.95\columnwidth]{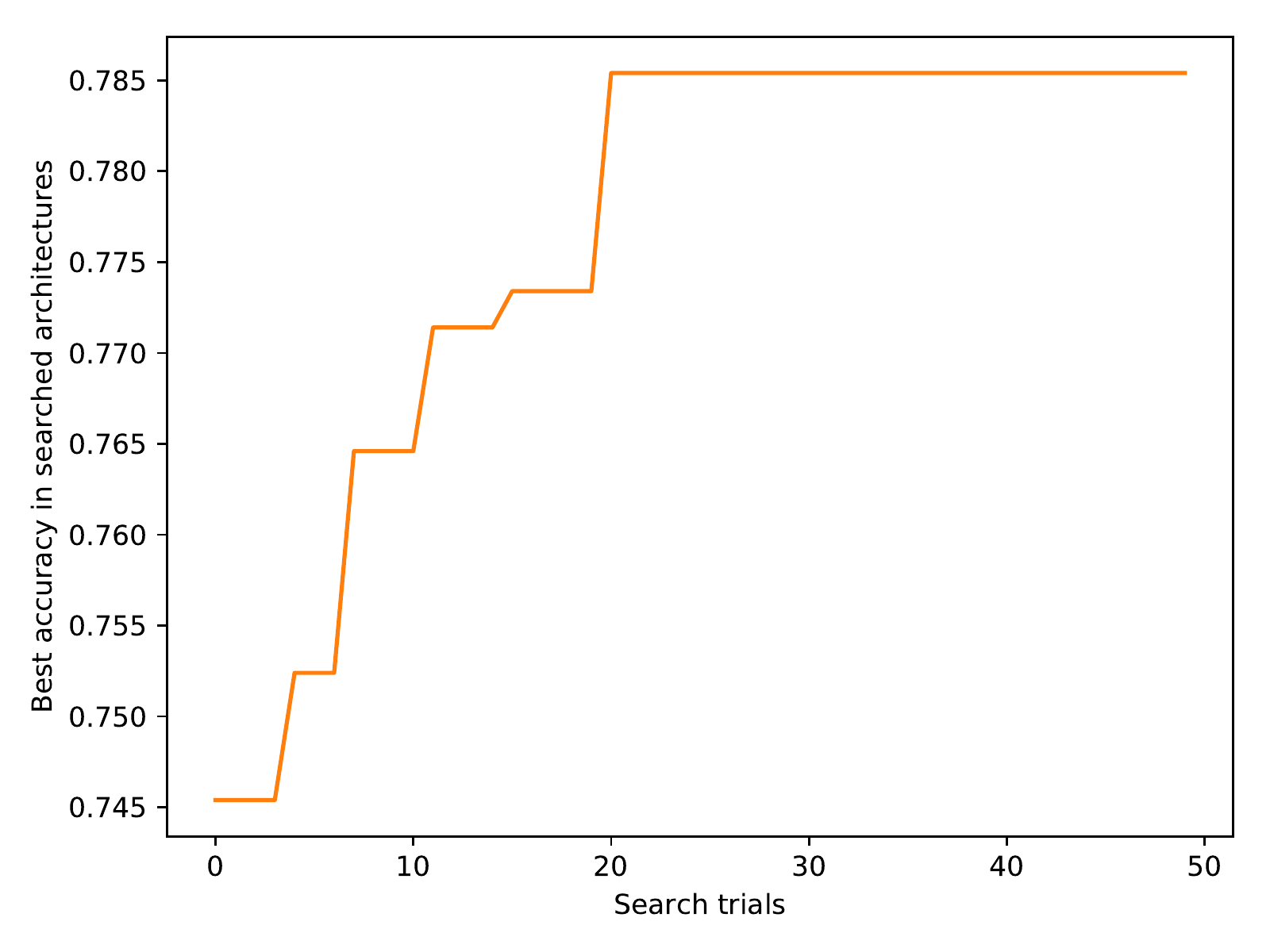}
    \caption{The best score after each search trial for CIFAR-100.}
    \label{fig:perf_step}
\end{figure}

Figure \ref{fig:perf_step} shows the best validation accuracies after each search trial for the CIFAR-100 dataset. We observed that after 20 iterations, the performance is no longer improved. We find that 50 trials is enough for reasonable performance on all the datasets that were analysed in this work.

\begin{table*}[htb]
\centering
\caption{Performance of automatic deep learning architecture search methods.}
\label{tab:comparison}
\resizebox{\textwidth}{!}{
\begin{tabular}{|l|l|l|l|l|l|l|l|}
\hline
\multirow{2}{*}{Method}                             & \multirow{2}{*}{Search} & \multirow{2}{*}{Parameter (CIFAR-10)} & \multirow{2}{*}{Finetune} & \multicolumn{4}{l|}{Error rate \% (single/ensemble)} \\ \cline{5-8} 
                                                    &                         &                                      &                           & CIFAR-10  & CIFAR-100   & SVHN      & FER2013        \\ \hline
Gastaldi \cite{gastaldi2017shake}                   & -                       & 26.2M                                & -                         & 2.72      & 15.85       & -         & -              \\ \hline
Zagoruyko, Komodakis \cite{zagoruyko2016wide}       & -                       & -                                    & -                         & -         & -           & 1.54      & -              \\ \hline
Tang \cite{tang2013deep}                            & -                       & -                                    & -                         & -         & -           & -         & 30.7           \\ \hline
Ours                                                & 50                      & 2.1M                                 & No                        & 5.06/4.09 & 21.60/17.48 & 2.84/2.34 & 28.28/25.49    \\ \hline
Zoph, Le \cite{zoph2016neural}                      & 12800                   & 7.1M                                 & Yes                       & 4.47      & -/-         & -/-       & -/-            \\ \hline
Baker et al. \cite{baker2016designing}              & 3500                    & 11.18M                               & Yes                       & 6.92/7.32 & 27.14/-     & 2.28/2.06 & -/-            \\ \hline
Real et al. \cite{real2017large}                    & 15000                   & 5.4M                                 & No                        & 5.40/4.40 & 23.00/-     & -/-       & -/-            \\ \hline
Miikkulainen et al. \cite{miikkulainen2017evolving} & 7200                    & -                                    & -                         & 7.30/-    & -/-         & -/-       & -/-            \\ \hline
Zoph et al. \cite{zoph2017learning}                 & 20000                   & 3.3M                                 & -                         & 3.41/-    & -/-         & -/-       & -/-            \\ \hline
Suganuma et al. \cite{suganuma2017genetic}          & 300                     & 1.68M                                & -                         & 5.98/-    & 23.47/-     & -/-       & -/-            \\ \hline
Cai et al. \cite{cai2017reinforcement}              & 480                     & 19.69M                               & -                         & 5.70/-    & -/-         & -/-       & -/-            \\ \hline
Negrinho et al. \cite{negrinho2017deeparchitect}    & 64                      & -                                    & -                         & 11.00/-   & -/-         & -/-       & -/-            \\ \hline
\end{tabular}
}
\end{table*}

Table~\ref{tab:comparison}\footnote{The actual number of architectures searched was not provided in Real et al.\cite{real2017large} The authors provided only wall-clock time. As they mentioned the population size is 1000 and also from Figure-1 of the paper, our estimated number of architectures searched can not be less than 15000.} shows the comparison between several automatic architecture search methods on different datasets along with the best performance found in the corresponding literature. 
From the results, we observe that our approach achieved competitive performance with respect to the number of searched architectures and the model complexity measure in the  number of parameters.
For CIFAR-10, it achieved better results than most of the methods except Zoph, Le \cite{zoph2016neural} and Zoph et al. \cite{zoph2017learning} where both the number of searches and the number of model parameters are much larger than our approach. For the CIFAR-100 dataset, our method achieved better results than all other architecture search methods. For the FER2013 dataset, it achieves state-of-the art classification accuracy. 
For the SVHN dataset, our approach achieves competitive performance compared to other methods, but with fewer model parameters. 

\begin{figure}[htb]
\includegraphics[width=0.95\columnwidth]{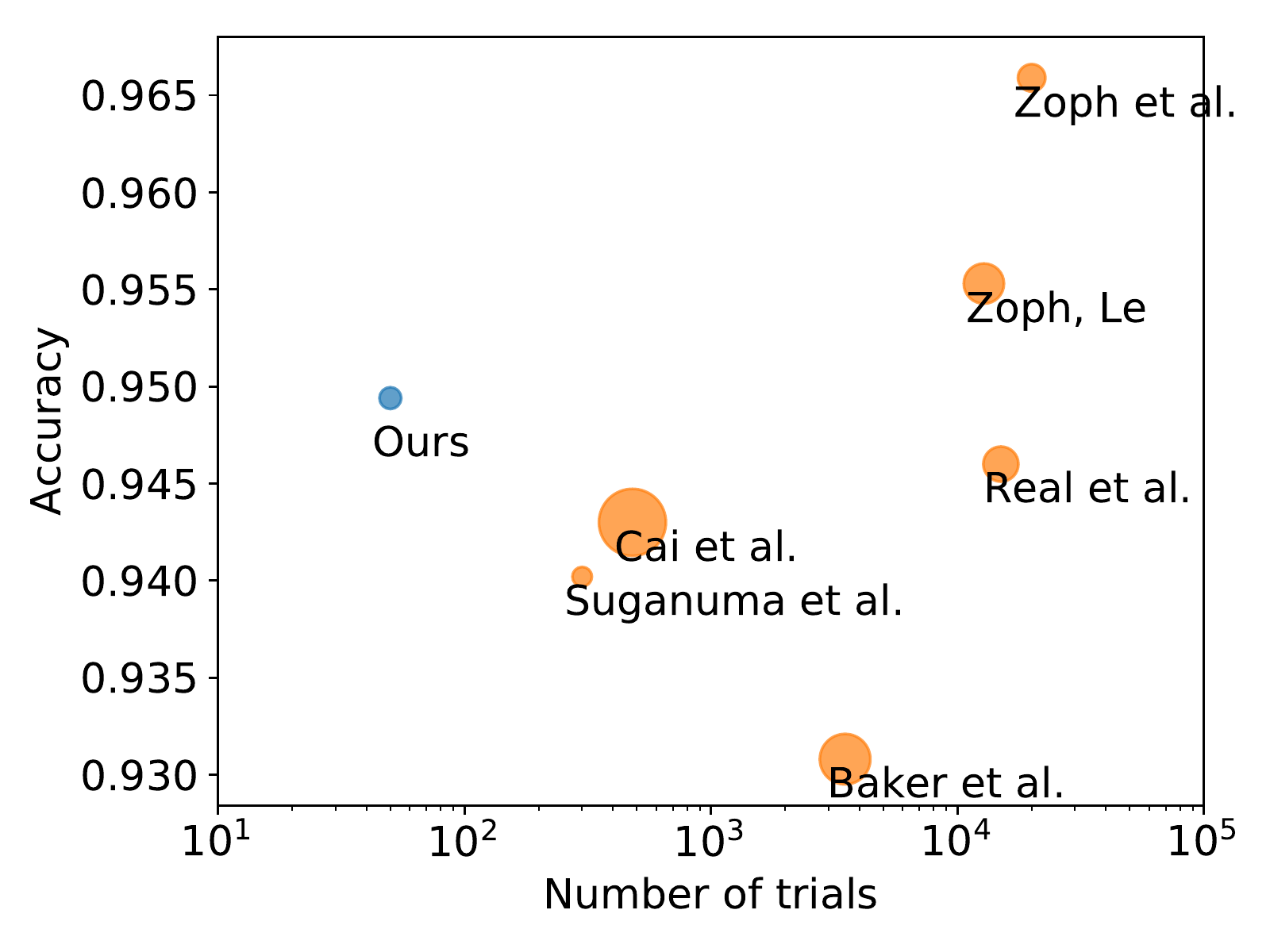}
    \caption{Performance comparison of different architecture search methods on CIFAR-10.  The size of circles corresponds to the complexity of the final \ac{cnn} model (see Table~\ref{tab:best_arch}).}
    \label{fig:perf}
\end{figure}

\section{Discussion}
Figure~\ref{fig:perf} visualizes the results of the different architecture search strategies on the CIFAR-10 dataset. For approaches that start from basic building blocks such as \cite{zoph2016neural,real2017large,zoph2017learning}, it is very hard to learn an effective block and replicate it into a very deep network in a short amount of time. These techniques take 3 orders of magnitude more trials than our approach to discover architectures that are only marginally better in terms of accuracy. 

The size of each data point in the figure represents the model complexity as measured by the number of weight parameters. The model discovered by our approach is smaller while still maintaining good prediction performance. This is due to the design of the search strategy that limits the number of branching options, the number of blocks, and enforces repeatable blocks to assemble the architecture. In effect we constrain our search to small networks of repeatable blocks instead of arbitrarily deep networks with unconstrained layer sizes.

The relatively small size of our models also has computational benefits. These models require fewer resources and can be trained faster than architecture that are discovered by extensive search. Although a smaller model may not be as representative as more complex ones, we are still able to achieve comparable performance. Finally, these models being smaller are also more generalizable and exhibit transferability. We designed a deep CNN using the best building block of CIFAR-10 experiment and trained that network on the CIFAR-100 dataset achieving an error rate of 21.8\%.

The best architecture for each dataset is presented in Table~\ref{tab:best_arch}.

\subsection{Ensemble of search results}
If the gains in reducing training time and model complexity are not enough to offset the reduction in performance, an effective way to improve results is to construct ensembles of searched models. Even though the top 10 models share similar performance, in our experiments, an ensemble prediction further improved accuracy.  

\begin{figure*}[t]
  \centering
  \includegraphics[width=0.96\textwidth]{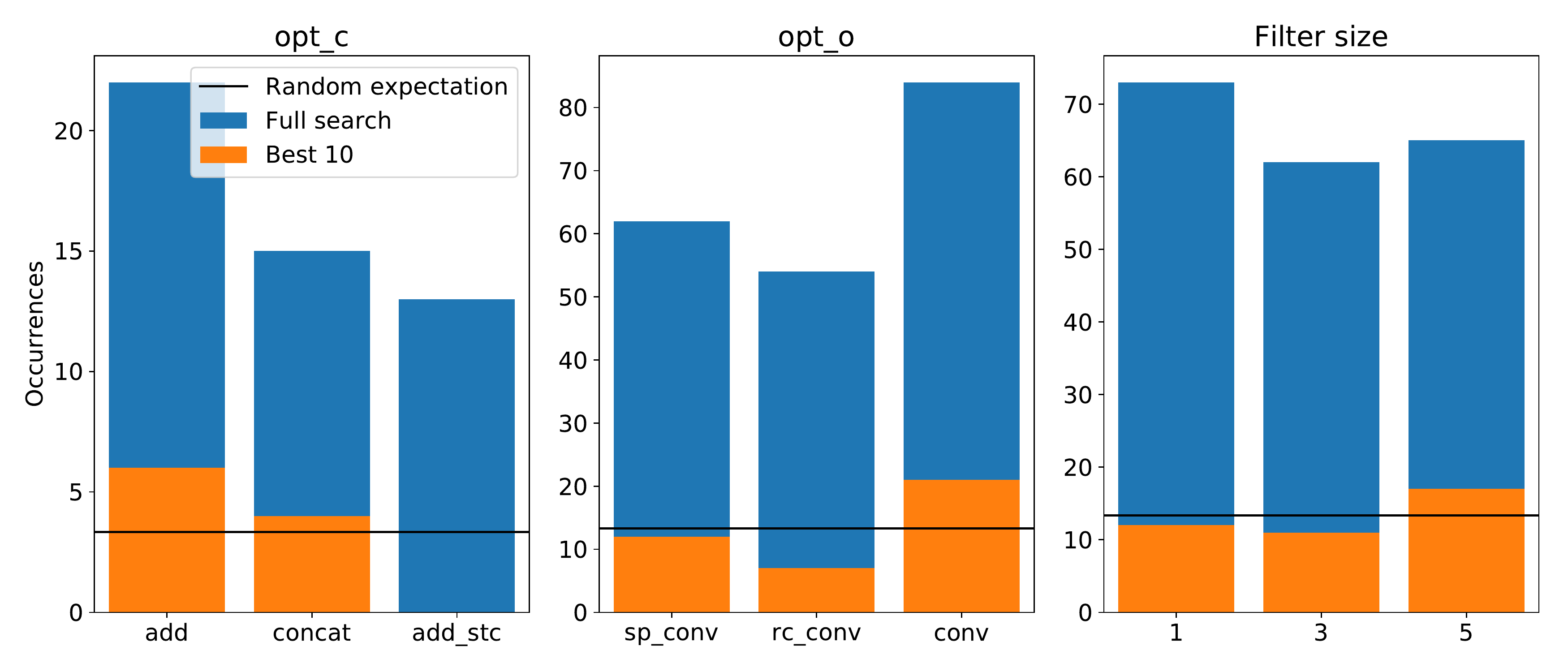}
  \caption{Histogram of building blocks of the searched deep \ac{cnn} architectures. Blue bars are from all the searched architectures, and the orange bars are for top 10 models that form the ensemble. The black lines are the expected occurrences for the top 10 models, if randomly selected.}
  \label{fig:emsemble}
\end{figure*}

We also compare the types of blocks used by the different models contrasting the blocks learned for all models vs those learned for the top-10 (used to construct the ensemble model). Figure~\ref{fig:emsemble} shows the histogram of occurrences of building blocks used in the models searched for the CIFAR-100 dataset. This analysis allows us to understand how the different block components contribute to the result and provides some hints for future \ac{cnn} design. For instance, in the \emph{opt\_c} block, \emph{add} and \emph{concat} operations are present in the models that form the ensemble (top 10 models) while \emph{add\_stc} is absent potentially showing that it is not a good choice to be a part of the architecture building block. We also see that both \emph{conv} and a filter size of 5 are slightly favored in our problem settings. However, this analysis is preliminary and as the number of searches are limited, we do not draw any firm conclusions. In addition, with more search iterations, optimization algorithms that leverage relations between hyperparameters may outperform the random search method used in this paper. 

\begin{table*}[]
\centering
\caption{Building blocks for best model with respect to validation set of each dataset.}
\label{tab:best_arch}
\begin{tabular}{|l|l|l|l|l|l|}
\hline
\multirow{2}{*}{Dataset} & \multicolumn{4}{l|}{Operation type} & \multirow{2}{*}{Combination  type}                                           \\ \cline{2-5}
                         & Branch-1                            & Branch-2           & Branch-3           & Branch-4           &               \\ \hline
CIFAR-10                  & \emph{conv(5)}                      & \emph{sp\_conv(1)} & \emph{sp\_conv(3)} & \emph{rc\_conv(3)} & \emph{add}    \\ \hline
CIFAR-100                 & \emph{conv(5)}                      & \emph{conv(1)}     & \emph{sp\_conv(3)} & \emph{sp\_conv(3)} & \emph{add}    \\ \hline
SVHN                     & \emph{conv(1)}                      & \emph{rc\_conv(3)} & \emph{conv(5)}     & \emph{rc\_conv(1)} & \emph{concat} \\ \hline
FER2013                  & \emph{conv(5)}                      & \emph{conv(3)}     & \emph{sp\_conv(5)} & \emph{rc\_conv(1)} & \emph{concat} \\ \hline
\end{tabular}
\end{table*}

\section{Conclusion}

An important advance that improves deep learning performance is the use of building blocks - small branching/spanning convolution blocks with pooling and batch normalization layers that can be repeated to construct deep architectures. However, finding an effective building block for a task essentially adds another set of parameters to an already rich hyperparameter space. This ever increasing search space for hyperparameters means that effective architecture design is often the result of extensive search combined with deep expertise that allows experienced modelers to restrict search to certain promising combinations of parameters. It is not too much of a stretch to say that effective architecture design is partly art rather than completely science. 

One way to even the playing field for architecture design is to allow architectures to be learned rather than explicitly designed. In this paper we have shown that posing the design of a building block as a search problem over a limited set of building block components is effective in generating CNNs for image recognition. The key takeaway of our approach is that simple choices are often as effective as more complicated approaches to architecture design with the added advantage of generating smaller models with better generalizability. We limit our search space to a small set of components chosen from existing designs, limit the number of building blocks to only four, and use random search to explore this space. All these choices simplify our search space and greatly reduce search time. What is remarkable is that our technique discovered architectures that are, in most cases, better in performance than existing architecture search approaches and perform close to state-of-the-art (and even improving state-of-the-art in one case). There is as always a trade-off in terms of performance vs cost and we believe for those cases where cost (in terms of model design/training time) is an important factor, our approach can provide solutions that are comparable in performance.

There remain many avenues for future work. We are interested in expanding this technique to non-vision datasets and model architectures, including recurrent networks and even reinforcement learning methodologies. In addition, we are also interested in better understanding how much transfer is possible between blocks learned via random search. Finally, there is still some work to understand exactly what this simplified approach captures about architecture design that allows it to be so effective. This understanding could pave the way to designing newer building block components that are learned from more primitive blocks without the need for large number of trials.

\bibliographystyle{ieee}
\bibliography{main-minimal}

\end{document}